\documentclass{article}

\usepackage[T2A]{fontenc}
\usepackage{placeins}

\usepackage{amssymb,amsfonts,amsmath}
\usepackage{cite,enumerate,float}
\usepackage{color}
\usepackage{tikz}
\usetikzlibrary{arrows,snakes,backgrounds}

\def\be{\begin{eqnarray}}
\def\ee{\end{eqnarray}}
\def\nn{\nonumber}

\def\p{\partial}

\def\s{\theta}

\definecolor{red}{rgb}{1,0,0}
\definecolor{orange}{rgb}{1,0.5,0}
\definecolor{violet}{rgb}{0.7,0,1}

\definecolor{dgreen}{RGB}{0, 120, 0}

\newcommand\hide[1]{}

\hoffset= - 1.0in         

\begin{document}

\hfill MIPT/TH-10/26

\hfill IITP/TH-10/26

\hfill ITEP/TH-10/26

\bigskip

\centerline{\Large{A new Uncertainty Principle in Machine Learning
}}

\bigskip

\bigskip

\centerline{\bf V.Dolotin and A.Morozov}

\bigskip

\bigskip

\centerline{\small {\it MIPT, Dolgoprudny, 141701, Russia}}
\centerline{\small {\it NRC ``Kurchatov Institute", 123182, Moscow, Russia}}
\centerline{\small {\it Institute for Information Transmission Problems, Moscow 127994, Russia}}
\centerline{\small {\it ITEP, B.Cheremushkinskaya, 25, 177259, Moscow, Russia}}

\bigskip

\bigskip

\centerline{ABSTRACT}

\bigskip

{\footnotesize
Many scientific problems in the context of machine learning can be reduced
to the search of polynomial answers in appropriate variables.
The Hevisidization of arbitrary polynomial is actually provided by one-and-the same
two-layer expression.
What prevents the use of this simple idea is the fatal degeneracy of the
Heaviside and sigmoid expansions, which traps the steepest-descent evolution
at the bottom of canyons, close to the starting point, but far from the desired
true minimum.
This problem is unavoidable and can be formulated as a peculiar uncertainty principle --
the sharper the minimum, the smoother the canyons.
It is a direct analogue of the usual one, which is the pertinent property
of the more familiar Fourier expansion.
Standard machine learning software fights with this problem empirically,
for example, by testing evolutions, originated at randomly distributed starting points and then selecting the best one. Surprisingly or not, phenomena and problems, encountered in ML application to science
are pure scientific and belong to physics, not to computer science.
On the other hand, they sound slightly different and shed new light on the
well-known phenomena -- for example, extend the uncertainty principle from Fourier
and, later, wavelet analysis to a new peculiar class of nearly singular
sigmoid functions.

}

\tableofcontents

\bigskip

\bigskip

\section{Introduction}

Application of machine learning (ML) \cite{ML,MLmath,MLphys,MLde,frugality,MPphysgeom,MLphyssci,Chervov}
to scientific problems is a challenge \cite{ML1,ML2}.
The method was designed to search for {\it associations}
in the absence of definite and well defined answer,
what is technically realized as a steepest descent in the landscape
with many false minima.
Scientific problems are absolutely the opposite --
they have the unique true answer, and the task is to find it,
and not something ``similar'' at an undefined distance from the truth.
In ML context this means that we need to find the {\it true} minimum
and {\it avoid} getting into the false ones.
This is a principal difference, which is not easy to overcome.\footnote{
It may seem that the problem is similar to that in {\it quantum computing},
where the true answer is found only with some {\it probability}.
Still again there is a difference -- in that case the wrong answers have no
similarity to the true one, they are clearly wrong, and this can be easily tested,
what makes {\it quantum computing} a justified way to solve NB problems,
where it is difficult to find the answer but easy to check if it is true.
In machine learning the problem is posed differently, and testing the answer
is not always simple.
What is worse, the probability to get the true answer is not controlled,
and it can easily be infinitesimally small, as will be explained in the text below.
In other words, the uncertainty principle not only makes learning difficult and expensive.
it puts into doubt the answer -- for educated system it can be easy to find,
but difficult to check, if it is the {\it true} one. }
The notion of "true minimum" needs a proper interpretation/specification, discussed in Section \ref{ambig}.
Still, as it often happens,
after tremendous success in its mainstream application,
the ML method starts being applied beyond it --
and we witness mounting attempts to use it in pure exact science.

A serious analysis of such applications should include several steps.

First goes the very possibility to formulate a scientific problem
in terms of a minimum of certain "potential".
Moreover, ML implies that this potential has a rather peculiar form --
a combination of iterated  Heaviside step-functions.
Therefore applicability of ML depends on the possibility of {\it Heavisidization} --
converting various scientific results into the minima conditions of such
strangely-looking potentials.

Second, such conversion naturally leads to additional degeneracies,
and one needs to control and handle the emerging {\it gauge invariance}.
Moreover, this is a kind of unavoidable problem --
which mounts to the level of a new {\it uncertainty principle}.
{\bf The more pronounced is the minimum of the functional,
the more uncertain is its Heaviside approximation} --
the more coefficients remain undefined in expansion through
Heaviside functions.
This sounds somewhat paradoxical -- as often happens to uncertainty principles --
but in fact is a rather simple and universal property of expansions
in functions with Heaviside asymptotes --
just like conventional uncertainty principle in quantum mechanics
is a simple property of Fourier expansions
(the sharper the function, the more is the number of harmonics in its expansion).
In fact, the abundancy of false minima and cyclic evolutions
is a source of inspiration for conventional ML,
but it is an unwanted {\it difficulty} for its scientific applications.

Third, being a computer method, ML requires (i) discretization of the
steepest descent together with the  (ii) smoothing of Heaviside functions,
substituting them by the celebrated sigmoids.
One could expect that at this step the false minima disappear,
but in fact this is not the case.
At best we can get smooth valleys/canyons with slightly lifted degeneracies
but a very long time needed to travel along their bottoms. Additional problems are caused by various (iii) boundary effects,
typical for the switch from continuous limit to finite approximations.

Fourth, the publicly available ML software like the {\it TensorFlow} package,
uses particular versions of \ ``training'', which deviate from the standard steepest descent:
instead of solving the evolution equations till a stable point is achieved,
they make just one step and then switch to another training samples sequence
(starts from a new point, not the one reached at the previous step).

\hide{This is equivalent to switching from one type of averaging to another,
very much like switching from time to phase space averages in thermodynamics, --
what is often, but not always the same.}
In practice this is the trick which saves the steepest descent evolution from being
trapped in a canyon:
it rather tries different paths and looks for a point where different canyons intersect
without slowly traveling along one of them towards the intersection.

The list can be continued, but
in this paper we do not go far beyond these particular steps.
Our goal is to illustrate the new {\bf ML uncertainty principle}
with the simplest example of scientific problem -- the study of polynomials.
All what is needed to Heavisidize {\it arbitrary polynomial of arbitrary
number of variables} is just a {\it two-layer} network --
and it is quite interesting to see how (and if?!) the entire algebraic geometry
can be reduced to the work with such networks and their TensorFlow realization.
We are of course far from exhaustively resolving this puzzle --
but hope to explain what it smells like
and hope to convince other people to think more on this kind of problems.

\bigskip

This paper is built on and heavily uses the introductory analysis of {\bf ML in science},
presented in \cite{ML1,ML2}.
We briefly remind the main points in s.\ref{remind}.
As already said, the main statement for the purposes of this paper is that {\it any} polynomial
of any number of variables can be represented by the
formula (\ref{Pollev2}), i.e. the {\bf two-layer network  should be sufficient to deal
with all polynomial problems}.
From this perspective the only difference between various problems with polynomial
answers can be (and is) in the way the training set is generated for the ML session with
{\bf one and the same two-layer formula (\ref{Pollev2}) and one and the same  two-layer ML network}.

The main focus of this paper is two-fold.

a) We explain the way a given polynomial of a single variable $x$ is brought
into the form (\ref{Pollev2}), i.e. "Heavisidization of a given polynomial",
and why just {\bf two} layers are {\it formally} sufficient for  this purpose.

b) We analyze the degeneracy problem and formulate the {\bf uncertainty principle},
explaining the difficulty to use this nice Heavisidization.

These contradictory statements explain the main intrigue in scientific applications of ML.
In this discussion we comment on the issues of discretization,
smoothing, false minima and, not the least,
peculiarities of the {\it TensorFlow} software.
The outcome is a sample program in the attachment to the main text,
which we find useful and illuminating.

\bigskip
For those ready to omit the explanations and proceed straight to the consequences,
there is a self-consistent presentation/discussion of the direct analogy between spectral decompositions and neural networks in Sec.\ref{approx}.

\section{Paradigma of ML}

ML is a fastly developing branch of science, with no dogmas and methodological restrictions --
whatever can work for concrete application is immediately used and incorporated.
Still, there is a basic construction, which is still in use in most modern modifications,
and it is what we will call ML in this note.
At this level it has not so much to do with {\it computer} science,
it is rather a subject from {\it theoretical physics},
and we will use physical terminology and motivations in what follows.

\subsection{The task and the method of ML  }

The basic problem of ML is to
look for a global  minimum of the {\it loss functional}
\be
{\cal L} = \sum_a \left({ Y}(\vec x_a) - { y}_a\right)^2
\label{lossfunc}
\ee
Given is the  {\it training set}, consisting of pairs
$(\vec x_a| {  y}_a)$ ({\it input} and {\it output}),
and the search is for the function $Y(\vec x)$.
In the ideal, this function is supposed to map the input to the output
and the estimate functional takes it minimal possible value zero.

\begin{itemize}

\item{}
ML usually looks for $Y(\vec x)$ in the peculiar form
of iterated smoothened Heaviside functions, known as {\it sigmoids}.
Thus the question reduces to adjusting the set of coefficients ${W}$ in
such ansatz.
In this paper we restrict the choice of ${ Y}(\vec x)$
to the {\it two-layer} expression
\be
{ Y}(\vec x) :=
\int_I { W}_2^I\cdot \sigma\left\{\int_J W_1^{IJ}\cdot
\sigma\left(\vec W_0^J \vec x+B_0^J\right) + B_1^I\right\} + { B}_2
\label{level2}
\ee
where ${B}$ can be either added to the set of adjustment parameters ${W}$
or kept constant.\footnote{
According to modern versions of Kolmogorov–Arnold representation theorem \cite{KA}
even a {\it single} layer is sufficient to approximate arbitrary function,
but then the number of network nodes remains essentially unknown.
We thank V.B.Anikeev for sending us an inspiring old review \cite{Strunkov} on this subject.
}

\item{}
Another principle of ML is that the minimum is searched for by the
steepest descent method (SDM),
i.e. by finding
the solutions ${\bf W}(t)$
to the equations
\be
\dot{\bf W} = -\frac{\p{\cal L}}{\p {\bf W}}
\ee

\item{}
The standard SDM would look for the
the large-$t$ asymptotic of the solution ${\bf W}(t)$,
but we will see that in practice, because of the {it canyon problem},
this can take too long.
A phenomenological way out of this difficulty is
to try steepest descents  ${\bf W}(t)$ with different and even random
initial conditions.

\end{itemize}

\noindent
In fact, 
there are no reasons to restrict consideration to just (\ref{lossfunc}).
If additional structures are defined on the set of variables,
like topology or metric,
one can exploit the notion of close points, up to adding derivative terms to the functional.
We, however, postpone discussion of such ``details''  to another occasion.

\subsection{The basic program}

To solve the ML problem, formulated in the previous subsection
we use a simple program attached to the electronic version of this paper
and, when needed, illustrate our statements by associated screenshots.
Shown are:

A) the input $x$, output $y$, the starting values of
optimization parameters $W_{0,1,2}(t=0)$ and $B_{0,1,2}(t=0)$ in (\ref{level2}),

B) the desired Heavisidized answer, in the form of expected values of $w$ and $b$

C) the current values of $W(t)$ and $B(t)$, which vary with time $t$.

D) the deformation (smoothing) of Heaviside function.

The program allows to use not only sigmoids (as in true ML),
but also to make other choices and examine, say, the wavelet decompositions -- which, however, are beyond the scope of the present paper.

Our task will be to apply ML algorithm to the simplest polynomial problems
and use the program to illustrate the emerging problems and the ways
they are resolved.

\subsection{Specifics of {\it scientific} problems}

If we apply ML to scientific problems,
we assume that the training set is consistent with some definite function $y(\vec x)$ --
the true answer to our question, i.e. the true law of nature.
The scientific task is to find exactly this function and not anything else.
This is different from the usual applications of ML,
where no concrete function is obliged to exist,
and the task is to find a kind of a {\it probability  distribution},
which adequately {\it approximates} the experimental data.

Of course, when the function $y(\vec x)$ does exist,
it {\it is} a solution to the steepest descent equations, i.e. can be an outcome of ML.
Still there are questions:

a) if it is the only one,  and

b) how easy is it to reach exactly this solution by the steepest descent?

To this one can add at least two extra questions,  related to actual

In addition ML actually uses a discretized version of (\ref{level2}),
where integrals are substituted by discrete sums
and indices $i,j$ take integer values.
This means that we deal with polynomials at integer/discrete points only, and associated ambiguities will be discussed as {\it discretization problems}.

Still another point is that ML uses the smoothed version of Heaviside functions, {\it sigmoids},
and the corresponding phenomena will appear under the  name of {\it smoothening problems}.

\section{Heavisidization of polynomials}

\subsection{Heaviside function}

We define  the Heaviside function as
\be
\theta(x)  = \left\{\begin{array}{ccc} 1 & {\rm for} & x>0 \\ \\
0 &{\rm for} & x\leq 0
\end{array}\right.
\label{thetadef}
\ee
Note that in this definition $\theta(0)=0$.
Then it possesses the projector property
\be
\s^{\circ 2}(x) :=\s\Big(\s(x)\Big) = \theta(x)
\label{proj}
\ee
Further, this function realizes logical operations {\bf AND} and {\bf OR}:
\be
\begin{array}{ccccc}
\wedge(a, b):=\s\Big(\s(a) + \s(b) - 1\Big) = 1  & \ \text{only if} \
&  \s(a)=1  & {\bf AND} &  \s(b) = 1
\\ \\
\vee(a,b):=\s\Big(\s(a) + \s(b)\Big) = 1 & <=>
&  \s(a)=1  & {\bf OR} &  \s(b)=1
\end{array}
\label{andor}
\ee

\subsection{Basic example}

Let us try to solve the Heavisidization problem for an identity function $y(x)=x$:
\be
\sum_I B_I\theta(W_I x + b_I)   = x
\ee
and its smoothed version
\be
\sum_I B_i\sigma(W_I x + b_I)   = x
\ee
Exact solution exists:
\be
x = \int_0^\infty \theta(x-y) dy  - \int_0^\infty \theta(-x-y) dy
= \int_{-\infty}^\infty {\rm sign}(y)\cdot \theta\Big({\rm sign}(y)(x-y)\Big),
\label{solidentity}
\ee
i.e.
\be
B_I = {\rm sign}(I), \ \ \ \ X_I={\rm sign}(I), \ \ \ \ b_I=-{\rm sign}(I)
\ee
the problem is to find it by the steepest-descent method.

\begin{picture}(100,155)(100,-10)
\put(300,5){
\put(-90,50){\vector(1,0){180}}
\put(0,0){\vector(0,1){100}}
\put(0,75){\line(1,-1){75}}
\put(0,25){\line(-1,-1){25}}
\put(85,55){\mbox{$y$}}
\put(5,80){\mbox{$x$}}
\put(-20,30){\mbox{$-x$}}
\put(0,75){\circle*{3}}
\put(0,25){\circle*{3}}
\put(5,95){\mbox{${\rm sign}(y)(x-y)$
}}
\put(-50,115){\mbox{${\bf {\rm sign}(y)\cdot theta\Big({\rm sign}(y)(x-y)\Big)}$}}
\linethickness{2pt}
\put(0,75){\line(1,0){25}}
\put(25,50){\line(1,0){55}}
\put(-90,50){\line(1,0){90}}
}
\end{picture}

The first part of the question, is

1) whether the solution (\ref{solidentity}) of this type is unique,

2) the second part is how efficient in finding it is the steepest-descent method

3) the third one is what are the main obstacles
and

4) the forth part concerns the ways to overcome them.

\subsection{Elementary operations \cite{ML2} (a short deviation into pure math)
\label{remind}}

Heaviside function allows to define all the elementary operations,
either on integers or on real numbers.

First, on integers:
\begin{itemize}
\item
$\delta_n(x)$: 1 iffy $x = n$ is defined similar to the  zero of $(x - n)$
\be
\delta_n(x) = \s(x-n+1)-\s(x-n)
\ee
\item
Identity
\be
x=I(x):=\sum_{i=0}^\infty\s(x - i) - \sum_{i=0}^\infty\s(-x - i)
=\sum_{i\in\mathbb{Z}}(\text{sign}\  {i})\cdot\s\Big((\text{sign}\ i)(x - i)\Big)
\label{eval_map}
\ee
\item Addition
\be
x+y = I(x)+I(y)
\label{add}
\ee
For positive integers addition looks simpler:
\be
x+y = I(x)+I(y)
= \sum_{i=0}^\infty \s(x-i) + \sum_{j=0}^\infty \s(y-j)
= \sum_{i,j=0}^\infty \vee\,(x-i, y-j)
\label{addpos}
\ee
\item  Subtraction
\be
x-y = I(x) - I(y)
\label{subtr}
\ee
\item Multiplication of positive integers
\be
x\cdot y = \sum_{i,j=0}^\infty \s\Big(\s(x-i)+\s(y-j) -1\Big)=
\sum_{i,j=0}^\infty \wedge\,(x-i, y-j)
\label{mult}
\ee
\item
Division
\be
x/y=\sum_{i,j}\wedge\!\left(x-i,\ \frac{\delta_j(y)}{j}\right)
\label{t_div}
\ee
\item
Square root and roots of other degrees:
\be
x^{1/\alpha} = \sum_{i=0}^\infty \s(x-i^\alpha)
\label{power}
\ee
\end{itemize}

Second, continuous counetrpart:
\begin{itemize}
\item
A root of any function:
\be
\text{zero of}\ f =
 \int_{dx}  x \frac{d}{dx}\theta\Big(f(x)\Big)
\label{zerof}
\ee
\item
Identity
\be
x=I(x):=\int_{i=0}^\infty  \s(x - i) -\int_{i=0}^\infty\s(-x - i)
=\int_{i=-\infty}^\infty (\text{sign}\  {i})\cdot\s\Big((\text{sign}\ i)(x - i)\Big)
\label{eval_map_cont}
\ee
\item Addition and subtraction
\be
x\pm y = I(x)\pm I(y)
\label{add_cont}
\ee
Again, for positives  addition looks simpler:
\be
x+y = I(x)+I(y)
=\int_{i=0}^\infty \s(x-i) + \int_{j=0}^\infty \s(y-j)
= \int_{i=0}^\infty \int_{j=0}^\infty \vee\,(x-i, y-j)
\label{addpos_cont}
\ee
\item Multiplication of positive integers
\be
x\cdot y = \int_{i=0}^\infty \int_{j=0}^\infty \s\Big(\s(x-i)+\s(y-j) -1\Big)
=\int_{i=0}^\infty\int_{j=0}^\infty \wedge\,(x-i, y-j)
\label{mult_cont}
\ee
\item
Square root and roots of other degrees:
\be
x^{1/\alpha} = \int_{i=0}^\infty \s(x-i^\alpha)
\label{power_cont}
\ee
\end{itemize}

In what follows we do not make any difference between integers and reals --
at the level of Heavisidisation the formulas are just the same.
The difference will emerge
when we start imposing computer restrictions and need to discretize
the continuous formulas.

\subsection{Polynomials and generic 2-layer network(s)
}

Immediate corollary of above formulas is that we can describe
{\bf arbitrary polynomial, of arbitrary degree and arbitrary number of variables}
by just the {\bf two-layer} Heaviside formula
\be
\boxed{
{\rm Pol}(\vec x) =y(\vec x):= \int_I w_2^I\cdot\theta\left(\int_J w_1^{IJ}\cdot
\theta(\vec w_0^J \vec x+b_0^J) + b_1^I\right)
}
\label{Pollev2}
\ee
with some $x$-independent coefficients $w$ and $b$.
It sounds unexpectedly simple, but this is the case:
any polynomials can be represented by just two-layer Heaviside formula.

For example,
\be
2x^2 + 5x +3 = 2\int_{i_1=0}^\infty \int_{i_2=0}^\infty
\theta\Big(\theta(x-i_1)+\theta(x-i_2)-1\Big)
+ 5\int_{i=0}^\infty \theta\Big(\theta(x-i)\Big) + 3\theta\Big(\theta(1)\Big)
\ee
i.e. the multi-index $I$ includes the pairs $i_1,i_2$,
as well as $i$ and $1$ in the last term and so on.
Notation will be a little more uniform for homogeneous polynomial
\be
2x^2 + 5xy+ 3y^2 = \int_{i_1=0}^\infty \int_{i_2=0}^\infty
\left\{2\theta\Big(\theta(x-i_1)+\theta(x-i_2)-1\Big)
 + 5\theta\Big(\theta(x-i_1)+\theta(y-i_2) - 1\Big)
+ \right.\nn \\ \left.
+ 3\theta\Big(\theta(y-i_1)+\theta(y-i_2) - 1\Big)\right\}
\ee
and multi-index $I$ now includes three copies of the pair $i_1,j_1$.

\subsection{Ambiguity of (\ref{Pollev2})
\label{ambig}
}

Taking into account studies of gauge invariance in \cite{ML2}, there rises a question if this representation is canonical.

Here we need to underline the specifics of training process with respect to analytic approach. Each real-world training deals with a finite set of (input,target) pairs - training dataset $\alpha$. This dataset may actually have an (a priori unknown) analytic representation, say, being a polynomial function $f$. But training on a finite subset $\Gamma_\alpha\subset\Gamma=\text{Graph}(f)$ of the graph of such function (of "complete"\ dataset) makes the network to evolve to a state minimizing the loss functional $\cal L_\alpha$ specific for dataset $\alpha$. Even more, there may be various stable network states (local minimums) for which there is no formalizable way to prefer any of them, training based on gradient descent may tend to any of those states as far as only samples from $\Gamma_\alpha$ are used.
To illustrate, Fig.1 shows the result of training evolution of network parameters for $3\times 3$-determinant model, where the training input was in the space of matrix coefficients and the target values were the values of the corresponding determinant. Those results are essentially different depending in the initial values, while the training sets where the same in both cases.

\begin{figure}[h]
\begin{center}
\includegraphics[scale=0.8]{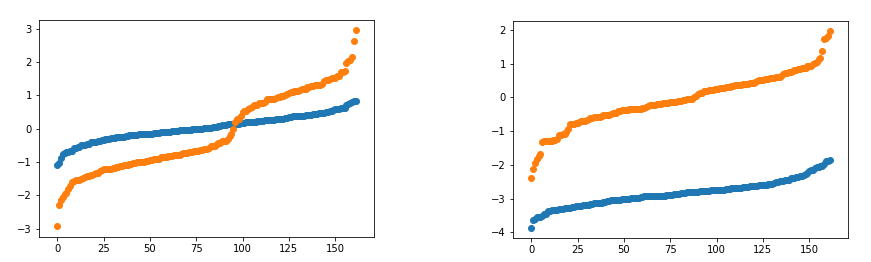}
\caption{Two different stable network parameters states (shown are about 150 ordered values), as the results of training for the same dataset but with different initial values}
\end{center}
\label{fig:hint_lim}
\end{figure}

Now if we pick another dataset $\beta$ we get another set of local minimums of $\cal L_\beta$, which in general are different from those for $\cal L_\alpha$.
As an example, Fig.2
shows matching of 400 selected (dataset $\alpha$) target values (blue) of a $3\times 3$ determainant as a function of matrix coefficients, with the values (orange) returned by a trained network. Orange points practically cover the target blue ones. But then we take another 400 values (dataset $\beta$) and the predicted orange values show an apparent mismatch on Fig.3.

\begin{figure}[h]
\begin{center}
\includegraphics[scale=0.5]{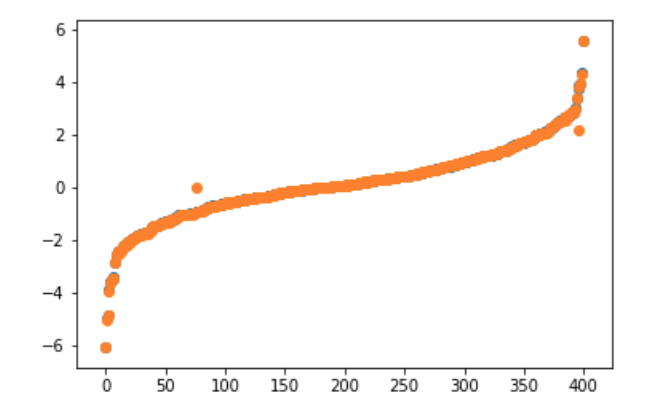}
\caption{Matching of 400 training target determinant values and a trained network output}
\end{center}
\label{fig:sample}
\end{figure}

\begin{figure}[h]
\begin{center}
\includegraphics[scale=0.5]{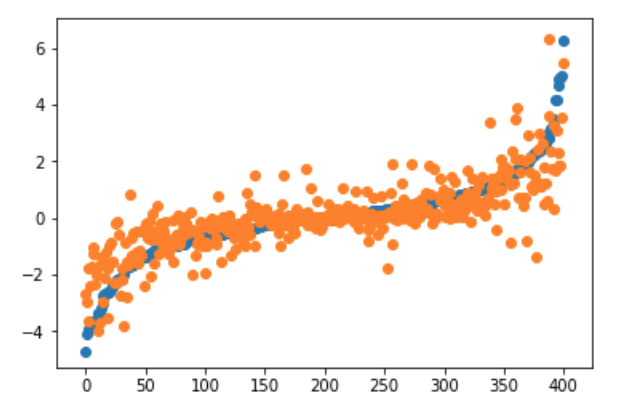}
\caption{Mismatch of the output returned by the same network, but taken for another set of 400 testing values}
\end{center}
\label{fig:test}
\end{figure}

Can we give a meaning to some {\it true} optimal network state, at least in case when the analytic answer $f(x)$ really exists? Here we can suggest a definition which reminds the "batches picking" used by training software to avoid trapping into canyons: for each descent step TensorFlow picks for computing the gradient some subset from the {\bf finite} dataset we gave it. So lets imagine that we have an unlimited time and instead of limiting ourselves to a particular dataset $\alpha$, we are enabled to pick more and more finite subsets belonging to the total {\bf infinite} graph $\Gamma$ of $f$, so that those subsets in the limit evenly cover the whole domain of $f$ and the descent steps are shared to each of those subsets in equal proportion. The trajectory in the space of network parameters for such a training is going to be pretty irregular and "long", but now we get a legal question:

\bigskip
{\it does this trajectory tend to some limit, or at least has accumulation point(s), as the time of above defined training procedure tends to $\infty$?}

\bigskip
At this stage we leave to the reader building (counter-)examples for this question/conjecture.

\section{
The origin of uncertainty principle}

\subsection{The difference from Fourier transform
}

The usual uncertainty principle is a pertinent property of Fourier transform.
At the level of linear transform
\be
F(x) = \sum_y G(y) K(x,y)
\label{lintrans}
\ee
one can wonder, how are related the properties of $F$ and $G$.

In Fourier case with the kernel $K = e^{\frac{ixy}{\hbar}}$ everyithing is simple:
\be
F(x) = 1         & \Longleftrightarrow &    G(y) = \delta(y) \nn \\
F(x) = \delta(x)     & \Longleftrightarrow &    G(y)=1
\ee
and, generically, the narrow $F$ are related to broad $G$ and vice versa,
the product of widths of $F$ and $G$ is $\sim \hbar$.

However, this is by no means a common property of all transforms (\ref{lintrans}).
For example for alternative kernel  $K(x,y) = \delta(x-y)$ we get and identity map $F(x)=G(x)$,
which does {\it not} change the widths of the functions.

Thus what we mean by uncertainty principle is a {\it complementarity} in a broader
sense of the word.

Heaviside transforms, used in ML, have another
important property --
they have extra parameters and are at risk of being degenerate.
This is exactly true only in extreme cases, but more generally we get
an approximate degeneracy: the speeds of the steepest descent are drastically
different in different directions, and one obtains a complicated relief,
where fast are only descents {\it to} the bottoms of ''canyons'',
while the descent {\it along} the canyon bottoms
can be very slow.
The analogue of uncertainty principle in this case states that {\bf the more
(training) parameters one introduces} --
thus the broader is the class of functions one can distinguish and recognize --
the more canyons one gets, thus {\bf the slower is the training and recognition processes}.

\subsection{Basic example}

We begin from the simplest possible example of identity function $y=x$ and try to
derive it by ML.
This will reveal immediate problems, which will be inherited by the more sophisticated examples.

\subsubsection{Degeneracy
}

So, our starting point is Heavisidization of identity function:
\be
y(x) = \int_0^\infty \theta(x-z) dx = x \ \ \ {\rm for} \ x>0
\label{idfun}
\ee
In ML problem we should find this function from minimization of
\be
{\cal L} = \int_{\text{domain in } x>0} \left(x - \int_0^\infty W\theta(wx+b-z) dz\right)^2 dx
\ee
with respect to parameters $W$, $w$ and $b$.

Let us first ignore $b$, i.e. put $b=0$.
Then we discover that
\be
x = \int_0^\infty W\theta(wx-z) dz
\ee
along entire valley $Ww=1$.
This demonstrates the ambiguity of Heavisidization:
getting in any point of the valley, we obtain a correct representation of the function $x$.
Despite we did not probably anticipate this in (\ref{idfun}), all these are correct answers,
more over they are easily obtained by ML.
If we pick up the simplest training sample -- all $x$ in between $0$ and $S$, then
\be
{\cal L} = \int_0^X   \left(x - \int_0^\infty W\theta(wx-z) dz\right)^2 dx =
\frac{X^3(Ww-1)^2}{3}
\ee
then the speed of the steepest descent to $W=w^{-1}$ is regulated by $X$:
\be
\dot W = -\frac{2X^3}{3}w (Ww-1) \nn \\
\dot w = - \frac{2X^3}{3}W (Ww-1)
\ee
and approximately
\be
Wv-1 = e^{-\xi t}
\ee
with
$ \xi = \frac{2X^3}{3}(W^2+w^2)$.
Actually $\xi$ is not quite constant, but fastly tends to one:
\be
\dot \xi \sim -\frac{2X^6}{3} Wv(Wv-1)\sim e^{-\xi t} \ \stackrel{t\rightarrow \infty}{\longrightarrow}\ 0
\ee

\subsubsection{Canyons
}

Once again, the ML-indpired answer is not unique in the sense that $W$ and $w$ are not defined
unambiguously, but all these answers are equivalent, and the problem is solved.
The difficulty arises, if we get stack at a point, which does {\it not} represent $x$,
i.e. if there are false minima.
But things are even worse -- even if there is no false minima,
there are canyons with a very slow descent to the point of minimum.

This happens in our elementary example when we revive the additional adjusting parameter $b$:
\be
y(x) = \int_0^\infty W\Big(\theta(wx-z) dz - b\Big) = W\cdot(wx -b)
\ee
Then (for the homogeneously distributed input $x$)
\be
{\cal L} =
\int_0^X\Big(x - W\cdot(wx -b)\Big)^2 dx = \frac{X(X^2w^2-3Xbw+3b^2)}{3}\cdot (W-x_+)(W-x_i)
\ee
with
\be
x_{\pm} = \frac{(2Xw - 3b \pm ib\sqrt{3})X}{2(X^2w^2-3Xbw+3b^2)}
\ee
has a very deep canyon, with the bottom at
\be
{\cal L}_{min} = \frac{b^2X^3}{4(X^2w^2 - 3Xbw + 3b^2)}
\ee
Note that at large $X$ the bottom is at the height $\sim X$, while the typical height of ${\cal L}$ is $\sim X^3$.
Thus the degeneracy for $b\neq 0$ is lifted, but the motion along the canyon towards its minimal point at $b=0$
is very slow -- as compared to the falloff into the canyon.

\subsubsection{Sigmoid}

With sigmoid the answer is no longer exact,
but this does not cause additional difficulties -- at least in our simple problem.

What happens when we substitute Heaviside function by sigmoid.
In this case
\be
\int_0^Z\frac{1 + \tanh\frac{x-z}{\Lambda}}{2}\ dz
= \frac{Z - \Lambda\log\cosh\frac{x-Z}{\Lambda} +\Lambda\log\cosh\frac{x}{\Lambda} }{2}
\ \stackrel{Z\longrightarrow \infty}{\longrightarrow} \
\frac{x + \Lambda\log(e^{x/\Lambda}+e^{-x/\Lambda})    )}{2} \nn \\
\ \stackrel{\Lambda\longrightarrow 0}{\longrightarrow}\
\left\{ \begin{array}{ccc} x && \frac{x}{\Lambda} > 0 \\ \\  0 &&\frac{x}{\Lambda}< 0
\end{array}\right.
\ee

\be
\int_0^Z W\frac{1 + \tanh\frac{wx-z}{\Lambda}}{2}\ dz
= W\frac{Z - \Lambda\log\cosh\frac{wx-Z}{\Lambda} +\Lambda\log\cosh\frac{wx}{\Lambda} }{2}
\ee

\subsubsection{Difference from wavelets
}

To demonstrate that canyons are {\it not just} a corollary of expansion in
overfilled basis, we consider an example of wavelet expansion, say,
of functions $F(x)$ in a linear combination of $e^{-\alpha(x-\beta)^2}$

We can look for wavelet expansion of $F(x)=e^{-x^2}$, by looking for the minimum of
\be
{\cal L}(\alpha,\beta)  = \int_{-\infty}^\infty \Big(e^{-x^2} - e^{-\alpha(x-\beta)^2}\Big)^2 dx
\sim 1 + \frac{1}{\sqrt{\alpha}} - \frac{2\sqrt{2}}{\sqrt{1+\alpha}}e^{-\frac{\alpha\beta^2}{1+\alpha}}
\ee
Clearly the minimum is at $\alpha=1$ and $\beta=0$
(in this example $F(x)$ can be matched exactly,  thus ${\cal L}_{min}=0$),
but deviation in both directions $\alpha$ and $\beta$
are nearly the same (the second derivatives in both directions are $\sim 1$).
This is a drastic difference from the canyon, where the second derivative along
is parametrically smaller than the one across.
The reason for this is the absence of degeneracy in any approximation.
The relevant {\it parameter}, which is the scale of the training domain $X$ also does not
show up in any significant way in the wavelet example.

\subsubsection{Moral
}

\begin{itemize}

\item{}One can wonder why can not we exclude the parameter $b$ (as a variant of reducing dimension), thus getting rid of the canyon problem.
The price to pay is a restriction of the form of the ansatz,
which jeopardize the possibility to represent some or even many functions.
This is exactly the point about the uncertainty principle:
{\bf if we want a broadly applicable ansatz/Heavisidization, we get the canyon problem}.
More concrete, parameter $b$ in above example is needed, if we want to Heaviside not only the identity function $y=x$,
but also $y=x+c$ with arbitrary $c$.
Then the degenerate solution is $w=W^{-1}, \ b=c\cdot W^{-1}$, and the deep canyon arises with the minimum at this point.
More generally, as we will see below, {\it any} polynomial can be Heavisidized by 2-level ansatz --
but all the $w,b$-parameters in it seem relevant for this possibility to exist.
And in result we can not get rid of canyons.

\item{}Are canyons always resulting from degeneracies
(nullifying some of the $w,b$ parameters, we get degenerate solutions).
Knowing this in particular cases, one could avoid/eliminate canyons.
But this is difficult to foresee/anticipate.

\end{itemize}

\section{Difficulties
}

Our simple two-level algorithms in section 2
actually suggest that the training process should converge to the simple
integer values of parameters $w$ and $b$, like $w = 0, 1$.
This proves the existence of solutions of appropriate type.
However, practical realization of training is not so simple.
There is a number of problems
which one needs to address and resolve.

The main one
is the emergence of degeneracies (valleys) in the estimate functional,
which can direct the steepest descent in a wrong direction.
Moreover, the valley problem is somehow enhanced when Heaviside functions
are changed for smooth sigmoids -- the vast Heaviside valleys get crossed by
additional narrow canyons.
At the same time this makes the valley
and the structures behind them
more pronounced
and provides new options to deal with the degeneracy
problems.

\subsection{Smoothing: from Heaviside to sigmoid
}

Heaviside function is an idealization of the conventionally used {\it sigmoid}:
\be
\sigma_K(x) = \frac{1}{1+e^{-Kx}}
\label{sigmadef}
\ee
which has simple hardware realizations.
In order to preserve to reasonable extent the projector  and other properties
of $\theta$, it often makes sense to consider a shifted sigmoid,
\be
\tilde\sigma_{K,\xi}(x) = \frac{1}{1+e^{-K(x-\xi)}}
\label{sigmadef}
\ee
with $K^{-1} \ll \xi \ll 1$.
A reasonable choice of $\xi$ requires some care,
in the first approximation one can assume that $\xi\sim K^{-1/2}$.

\subsubsection{The Heaviside valleys
and gauge invariance
}

Gauge invariance certainly does produce in the space of parameters directions tangent to level surfaces of the functional $L$, although to answer the question how much is that responsible for slowing down the training in the valleys, it needs to study the positioning of those directions in the neighborhood of minimums.

\subsubsection{A canyon
phenomenon
}

When we switch from Heaviside to sigmoid, a new phenomenon occurs.
For a smooth function $\sigma$ the estimate functional
\be
L(x) = |\sigma(w(\vec x))-b|^2
\ee
has a minimum along a whole hypersurface $w(\vec x)=b$,
which could never appear for Heaviside, where {\it this} $L$
takes just two values $|b|^2$ and $|1-b|^2$.
Whenever such continuous cannyon emerges,
numerical search for a minimum
is drastically slowed down -- the system can infinitely travel along the valley.

In reality the estimate functional is more complicated,
it contains a whole sum like
\be
L(x) = \sum_i |\sigma(w(\vec x_i))-b_i|^2
\ee
and the true minimum lies at the intersection
\be
\bigcap_i \Big\{w(\vec x_i)=b_i\Big\}
\ee
Still in most practical examples only one $x_i$ dominates,
while the others describe relatively small potentials
along the canyon
-- often smaller by the orders of magnitude.
This provides an end to the search along the valley,
but the time it takes can be very large.
Moreover, usually there is an entire hierarchy of valleys --
in the dominant approximation there are many flat directions,
in the next approximation degeneracy is lifted in just one of them,
after that in another and so on.

\bigskip

{\bf Example:
}

$$
h := x-> sum(w3[j]*s(sum(w2[j, i]*s(w1[i]*x+b1[i]), i = 1 .. N[1])+b2[j]), j = 1 .. N[2])+b3;
$$
$$
L := sum((h(tx[k])-ty[k])^2, k = 1 .. NT);
$$

Below -- a change of notation to $x=w^2_{[11]}$, $y=w^2_{12}$
{\footnotesize
\be
L =
(1/(1. + exp(-1.343852919*x + 10.40000000)) + 1/(1. + exp(-1.343852919*y + 10.40000000)) - 0.0002333160548)^2 +
\nn \\
+ (1/(1. +exp(-4.000000000*x + 10.40000000)) + 1/(1. + exp(-4.000000000*y + 10.40000000)) - 0.003317602160)^2 +
\nn \\
+ (1/(1. +exp(-6.656147081*x + 10.40000000)) + 1/(1. + exp(-6.656147081*y + 10.40000000)) - 0.04623154732)^2 +
\nn \\
+(1/(1. +exp(-7.686674218*x + 10.40000000)) + 1/(1. + exp(-7.686674218*y + 10.40000000)) - 0.1243831929)^2 +
\nn \\
+ (1/(1. +exp(-7.934699431*x + 10.40000000)) + 1/(1. + exp(-7.934699431*y + 10.40000000)) - 0.1566536472)^2
\nn
\ee
}
and one can easily plot the five items of the sum to see the hierarchy
(though in this example there are five items for just two directions $x$ and $y$).

The canyon phenomena is important for the following reason.
If we try to obtain $W=1$ from the condition $W\theta(w)=1$,
and substitute Heaviside $\theta$ by sigmoid $\sigma$ in the minimization process for
$L = |W\sigma(w)-1|^2$,
then instead of immediately finding $W=1$, the program starts travelling along
the valley for $w$ for a given $W\neq 1$, and refuses to adjust $W$ to $1$.

\subsection{From sigmoid to Heaviside}

Take the combination of sigmoids
$$h(a,b):=\sigma(a)+\sigma(b)$$

For a chosen point $(a_0,b_0)$ the function
$$L(a,b;a_0,b_0)=(h(a,b)-h(a_0, b_0))^2$$
clearly has its minimum at $(a,b)=(a_0,b_0)$, as well as at the whole 1-parametric set $\gamma$ of points given by the equation
$$\sigma(a)+\sigma(b)=h_0:=h(a_0,b_0)$$

For the sigmoid $\sigma(x)=\frac{1}{1+\exp^{-K(x-\xi)}}$ when we increase the value of $K$ and go to the limit $K\to\infty$ the sigmoid tents to $\s(x)$ in the space of generalized functions. At the same time the location of $\gamma$ tends to a singular set being the union of 2 rays:
$$\{a>1,\ b=1\}\cup\{a=1,\ b>1\}$$

At the same time for any $(a_0,b_0)\in\mathbb{R_+}\times\mathbb{R_+}$ the plot of $L(a,b)$ tends (independtly of $a_0, b_0$) to the step function on the plane being the plot of $(\s(a)+\s(b))^2$ (see Fig.4).

\begin{figure}
\begin{center}
\includegraphics[scale=0.35]{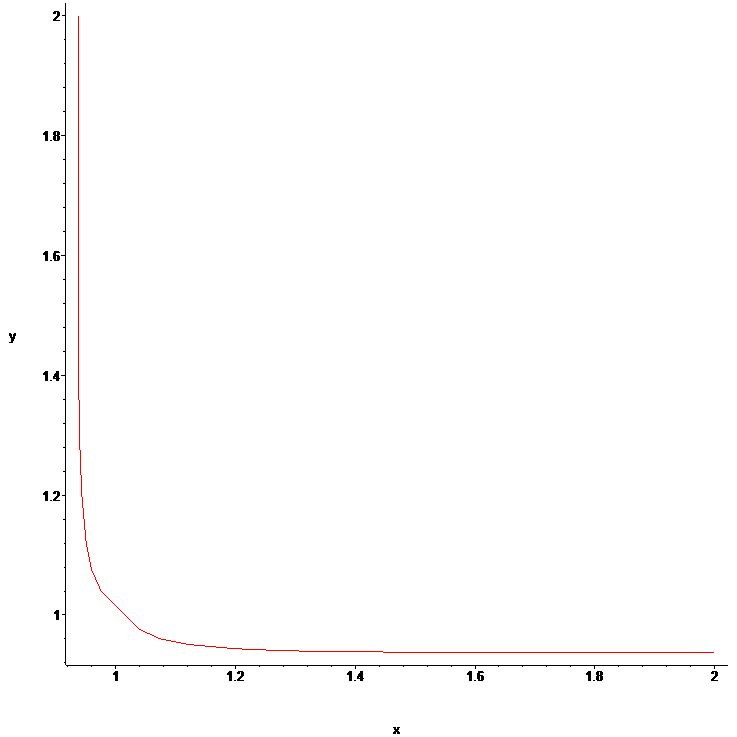}
\caption{The minimal curve $\gamma$ for $L(a,b;1,1)$}
\end{center}
\label{fig:1}
\end{figure}

\begin{figure}
\begin{center}
\includegraphics[scale=0.35]{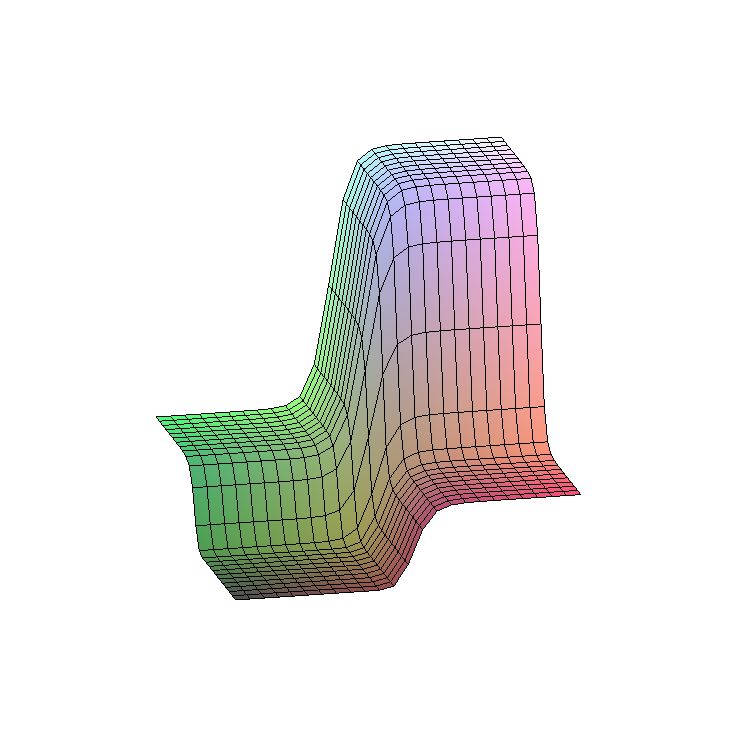}
\caption{The look of the plot of $L(a,b)$ in the region  $a\in[-0.5,3.5],\ b\in[-0.5,3.5]$ for $K=12$.}
\end{center}
\label{fig:2}
\end{figure}

\hide{
\begin{figure}
\begin{center}
\includegraphics[scale=0.35]{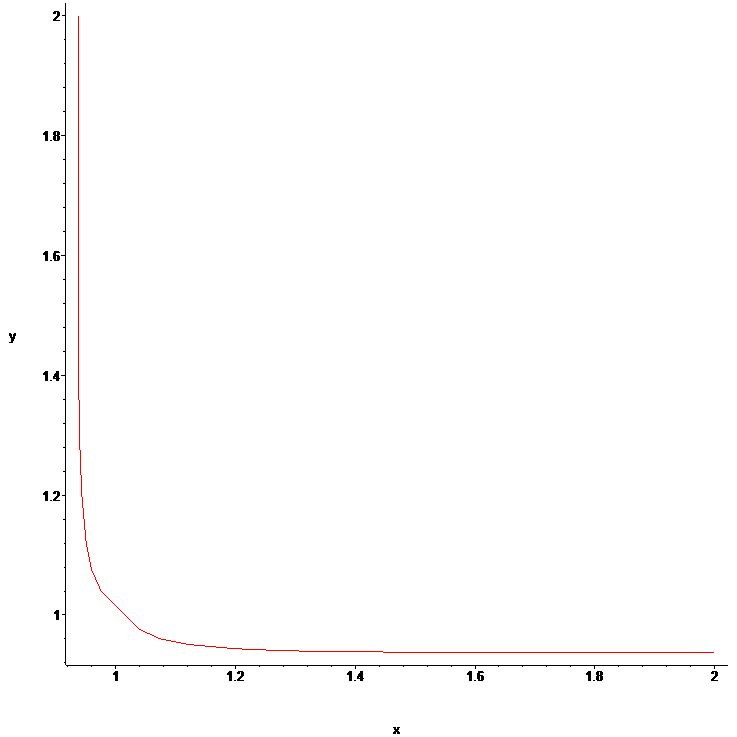}
\caption{The minimal curve $\gamma$ for $L(a,b;1,1)$}
\end{center}
\label{fig:1}
\end{figure}

\begin{figure}
\begin{center}
\includegraphics[scale=0.35]{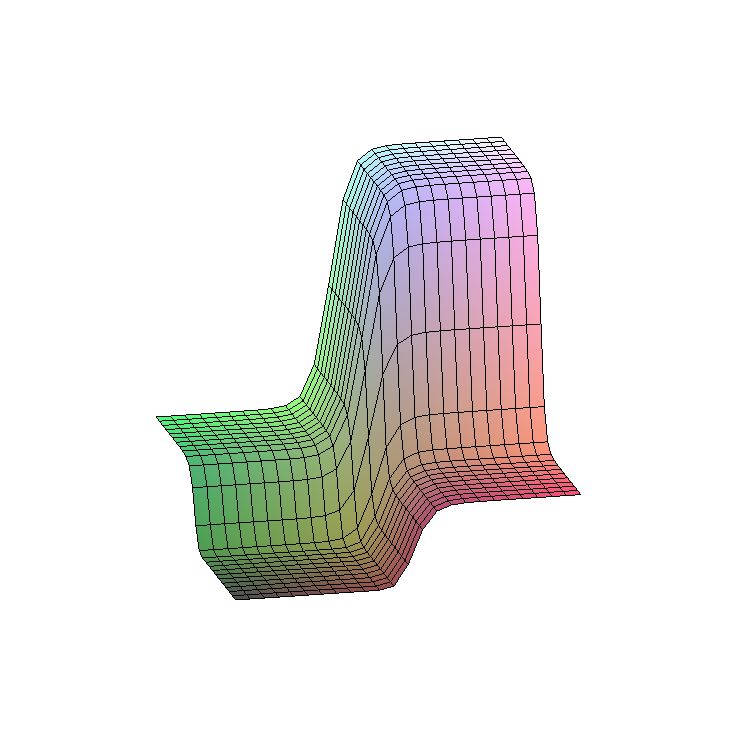}
\caption{The look of the plot of $L(a,b)$ in the region  $a\in[-0.5,3.5],\ b\in[-0.5,3.5]$ for $K=12$.}
\end{center}
\label{fig:2}
\end{figure}
}

\section{Analytical vs the TensorFlow approach
}

\subsection{Identity map}

\subsubsection{Steepest descent}

Let us begin from identity map
\be
a= I(a) = \int_{z=0}^\infty  \theta(a-z) dz
\ee
Heavisidisation at th r.h.s. implies the single-layer approximation
\be
{\cal I}_{W,B,w,b}(a) := B+\int_{dz'} W(z')\theta\left(b(z')+a\cdot \int_{z''=0}^\infty w(z',z'')  \right)
\ee
and then we need to apply the steepest descent method to
\be
{\cal L} =\frac{1}{2}\Big(A - {\cal I}(A)\Big)^2
\ee
In fact one can make a change of variables, so that the argument of $\theta$ becomes just $a-z$
with $z = -b(z)/\int_{z''=0}^\infty w(z,z'')$, and the caenge in the measure $dz'\longrightarrow dz$
is absorbed into rescaling of $W$.
Then actually
\be
{\cal I}(a) = B - \int_{z=0}^\infty W(z)\theta(a-z) dz
\ee
and we have a pair of steepest descent equations equation
\be
\dot B = -\Big(A - {\cal I}(A)\Big) \nn \\
\dot W(z) =  \theta(A-z)\cdot \Big(A - {\cal I}(A)\Big)
\ee
Clearly, there is a single stable point
\be
A =B+\int_{z=0}^\infty W(z)\theta(A-z) dz
\ee
Taking the $A$-derivative, we get $W(z)=1$ and then $B=0$.

\subsubsection{Variation of the theme
}

\be
{\cal L}= \frac{1}{2}\int \Big(y(x)-{\rm anzatz}(x|\alpha)\Big)^2 d\mu(x)
\ee
e.g.
\be
{\rm anzatz}(x|\alpha) = \int \theta(x-z) dz
\longrightarrow \left\{\begin{array}{ccc}
\theta\left(x-\frac{z}{\alpha}\right) \\
\theta\Big(x-z-\alpha\Big) \\
\theta\Big(x-z-\alpha\cdot{\rm sign}(z-z_0)\Big)
\end{array}
\right.
\ee
For the first option and $y(x)=x$
\be
\dot\alpha =  -(1-\alpha)\int x^2d\mu(x)
\ee
Instead of averaging over $x$ in ${\cal L}$ one can change  $x$ at every step,
i.e. substitute $x(t)$
and get
\be
\dot\alpha = -(1-\alpha) x^2(t)
\ee
This looks a little more arbitrary procedure, still it should also converge,
unless one takes special measures to prevent this.

\subsubsection{TensorFlow approach}

The first obvious difference when we switch to training via TensorFlow is
the smoothing of Heaviside function $\theta\longrightarrow \sigma$,
which makes above transformations illegitimate.
But this is not the main point.
Instead of solving the steepest-descent equations, TensorFlow uses a strategy of picking a subset of training samples for each descent step.

\subsection{Determinant
}

\subsubsection{$1\times 1$}

For $1\times 1$ matrix the determinant formula
becomes

\be
a = \int\s(\s(a - z))dz = \int\s(a-z)dz = {\cal I}(a)
\label{det11}
\ee

where we used the property $\s\circ\s = \s$.

\subsubsection{Concrete example
}

\be
\left\{D - \sum_{i,j=1}^\infty (C_1 \theta\Big( \theta(a-i)+\theta(d-j)-1\Big)
+C_2 \theta\Big( \theta(b-i)+\theta(c-j)-1)  \Big)\right\}^2
\ee

For determinant $m\times m$
\be
\sum_{\mu_1\leq\mu_2\leq...\leq \mu_n} C_{\mu_1\ldots\mu_n}
\int_{i_1...i_n=1}^\infty \theta\Big(  \theta(a_{\mu_1}-i_1)+ \ldots + \theta(a_{\mu_n} -i_n)+ n-1\Big)
\label{hdet}
\ee
\begin{itemize}
\item{
We want to see} that actually
$\mu_1< \mu_2 < …  $ (rather than $\leq$), and, further, all  $C=\pm 1$.

\item{Then we want to see} that it will work for $n\geq m$ and will not -- for $n<m$.
Will not work means that the answer will be unstable under the change of the training data.

\item{We want to see} what will change for smoothed $\theta\longrightarrow\sigma$ and for
discretization of the integral -- if the limit of large $N$ is somehow non-naive.

\item{We want to see} what happens if the training sequence is wrong (contains mistakes),
and what depends on the quantity/density of these mistakes.

\item{We want to see} what happens if we unjustly restrict $C$ -- say, fix some of them equal to $\pm 2$
instead of the true $\pm 1$.

\end{itemize}

\subsection{Towards non-linear algebra and knots}

The possibilities and the power of ML method should be tested on the problems
of non-linear algebra \cite{NLA,GKZ,MSh,Mandelbrot} --
a far-going generalization of determinant calculus to that of resultants and discriminants.
Its advantage is that it leaves us within the context of polynomial functions,
and at the same time asks questions about them, which are difficult to answer by usual methods.
This is a natural field for ML to compete, and hopefully it will one day attract attention,
which it definitely deserves.

Another promising test site is knot theory \cite{knots,KhR}.
Here some first attempts are being made \cite{GHRS,knotML} --
unfortunately, not yet within the systematic heavisidization program.

For a consideration of a closely related direction (polylogarithms) see \cite{DSZh}.

Of course, string theory \cite{UFN2} provides us with many other potential challenges,
still these two -- NLA and knots -- seem to be the first in the line.

\section{Numeric examples}
\subsection{Determinant $1\times 1$}

\subsubsection{Selecting the problem with known Heavisidisation}

In this particular case we actually deal with the identity map and
identity $\theta\circ\theta = \theta$
implies that the second
Heaviside function (second layer) is not really needed.
Still we keep it at some stages to keep the presentation of different determinants uniform.
So, what we want is to Maschine Learn the formula, which is the discrete version of (\ref{det11}):
\be
\sum_{i=0}^{M-1}\s(\s(x - i)) = \sum_{i=0}^{M-1}\s(x - i)
\label{heavanswer}
\ee

\subsubsection{Network configuration}

Associated network configuration is given by
\be
y(x)=\sum_{j=1}^Mw_2^{1j}\s\left(\sum_{i=1}^M w_1^{ji}\s(w_0^{i1}x + b_0^i)+b_1^j\right)
\ \stackrel{?}{\longrightarrow} \ \sum_{j=1}^{M}\s(\s(x - j)) = \sum_{j=1}^{M}\s(x - j)
\label{det11disc}
\ee
which has 1D input and output, and $M$ nodes on each of levels 1 and 2.

\subsubsection{Checking consistency}

Next, we check that the Heaviside answer at the r.h.s. of (\ref{det11disc})
is in the orbit of our ansatz at the l.h.s.
Indeed, it is:
\be
w_0^{i1} = 1 & j=1,\ldots, M \nn \\
b_0^j = - (j-1) &   j = 1,\ldots, M  \nn \\
w_1^{ji} = \delta_{ji}&   j,i. = 1,\ldots, M  \nn \\
b_1^j = 0 & j = 1,\ldots, M  \nn \\
w_2^{1j} = 1 &  j = 1,\ldots, M
\label{wbchoice}
\ee
i.e.
\be
y_0(x)=\sum_{j=1}^M \s\left( \s( x - j-1) \right)=\sum_{i=0}^{M-1}\s(\s(x - i))
=\sum_{i=0}^{M-1}\s(x - i)=x
\label{det11disc1}
\ee

\subsubsection{
A square}

It is instructive to compare this representation of the function $y(x)=x$
with a similar one for $y(x)=x^2$ for the particular value of $M=4$ which is a full square:
\be
y(x) = \s\Big(\s(x)\Big) + \s\Big(\s(x-1)\Big)+ \s\Big(\s(x-2)\Big) + \s\Big(\s(x-3)\Big) \approx x
\ee
\be
y(x) = \s\Big(\s(x) + \s(x)-1\Big) +  \s\Big(\s(x) + \s(x-1)-1\Big)
+  \s\Big(\s(x-1) + \s(x)-1\Big) +  \s\Big(\s(x-1) + \s(x-1)-1\Big) \approx x^2
\ee
In the first case $w_1^{ji} = \delta_{ji}$, i.e.
$w_1  = \left(\begin{array}{cccc}
1&0&0&0 \\ 0&1&0&0 \\ 0&0&1&0 \\ 0&0&0&1 \end{array}\right)$
and   $b_0^j = -(j-1)$.

In the second case $w_1  = \left(\begin{array}{cccc}
1&1&0&0 \\ 1&0&1&0 \\ 0&1&0&1 \\ 0&0&1&1 \end{array}\right)$
and $b_0^1 = b_0^2=0 , b_0^3=b_0^4=-1$
Moreover there is a number of alternative realizations
(which, break the symmetry of the matrix $w$ --
they naturally belong to the symmetry of $b_0$, i.e. a sort of $SU(2)^3$).

The open question: what are the formulas for $w$ and $b$ for generic $M$
and for generic polynomial.

\subsubsection{Artifacts of smoothing
\label{artif}}

Now we face the first serious challenge.
We need to substitute Heaviside functions (\ref{thetadef}) by their smoothed counterparts --
sigmoids (\ref{sigmadef}):
\be
\sigma(x) = \frac{1}{1+e^{-K^2(x-\xi)}}
\label{sigmadef1}
\ee
Some care is needed at this step to have (\ref{det11disc1}) preserved {\bf even at integer points  $x$}.
This is clear from the pictures for
$\sigma(x)-x$ for various choices of $\xi$.
In (\ref{thetadef}) we assume that it has exactly two zeroes at $a=0$ and $a=1$,
but in practice the curve is more complicated and can cross the real line
at one, two and three points:

The dependency of $\sigma(x)-x$, plotted as

\begin{verbatim}
K := 3;
xi := alpha/K;
sigma := 1/(1 + exp(-K^2*(x - xi)));
plot(sigma - x, x = -1 .. 1.1);
\end{verbatim}
\noindent
on the choice of $\xi = \alpha/K$ for various $\alpha$ is shown on Fig.6

\begin{figure}[H]
\begin{center}
\includegraphics[scale=0.45]{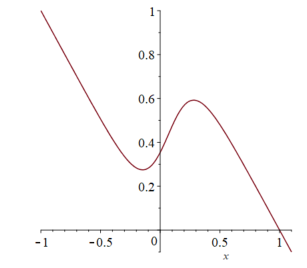}
\includegraphics[scale=0.45]{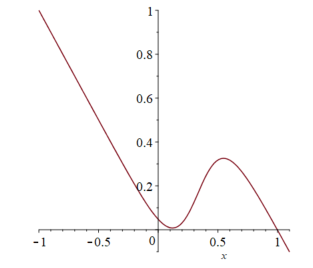}
\includegraphics[scale=0.45]{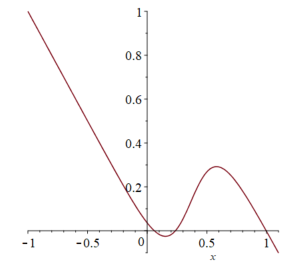}
\caption{$\sigma(x)-x$ for $\alpha=0.2,\ 1.0,\ 1.1$}
\end{center}
\label{fig:xi_alpha}
\end{figure}

What we need is the choice when the first two points are very close,
in practice the reasonable choice is $\xi\approx K^{-1}$,
but, as we see from above pictures, the coefficient also matters.

The values of
\be
L = \left( x-\sum_{i=0}^{10} \sigma(x-i)\right)^2
\ee
plotted as
\begin{verbatim}
K := 3;
xi := alpha/K;
L := add(1/(1 + exp(-K^2*(x - a - xi))), a = 0 .. 10);
plot((L - x)^2, x = -0.4 .. 11.4);
\end{verbatim}
for $\alpha=\pm 1$ are shown on Fig.7

\begin{figure}[H]
\begin{center}
\includegraphics[scale=0.65]{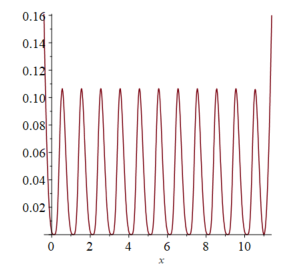}
\includegraphics[scale=0.65]{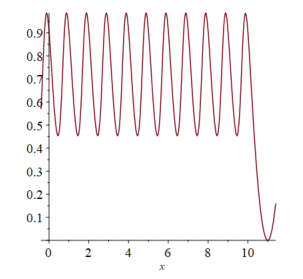}
\caption{dependency on the sign of $\xi$}
\end{center}
\label{fig:sig_sign}
\end{figure}

Additional care is needed when we switch from integers to rationals:
in the expression
$$\sigma_\epsilon(x) = \epsilon/(1 + \exp(-K^2(-a\epsilon + x - \xi))$$
\noindent
$\xi$ should also be adjusted to diminishing $\epsilon$.

The corresponding plots

\begin{verbatim}
K := 4;
xi := 1/K;
epsilon := 1;
L := add(epsilon/(1 + exp(-K^2*(-a*epsilon + x - xi))), a = 0 .. 10);
plot(L - x, x = -0.4 .. 11.4*epsilon);
\end{verbatim}
\noindent
for various $\epsilon$ are shown on Fig.8

\begin{figure}[H]
\begin{center}
\includegraphics[scale=0.55]{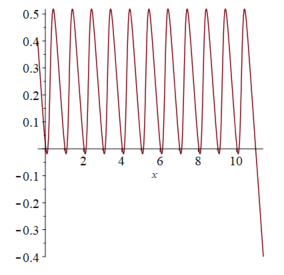}
\includegraphics[scale=0.55]{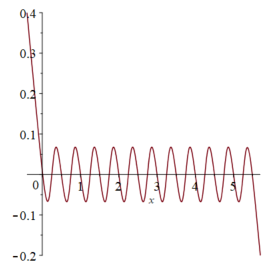}
\includegraphics[scale=0.55]{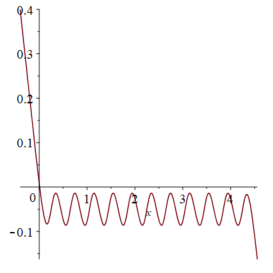}
\caption{dependency on $\epsilon=1,\ 0.5,\ 0.4$}
\end{center}
\label{fig:sig_eps}
\end{figure}

To summarize, when smoothing $\theta$ we should check that the desired solution
is still belonging to the orbit, and for this

$\bullet$ sigmoid should be appropriately centered ($\xi$ chosen carefully) and

$\bullet$ continuous limit $\epsilon\longrightarrow 0$, if needed, can also affect $\xi$

Instead of shifting sigmoid by $\xi$ we could shift the answer, i.e.
minimize/extremize the functional
\be
L_{\{Y_s,a_s\}}(w_1, b_1) = \sum_s\left(Y_s+\eta-\sum_{j=1}^{M}
\sigma\left(\sum_{i=1}^{M}w_1^{ji}\sigma(a_s-i-1)+b_1^{j}\right)\right)^2
\ee
with $\eta$ instead of $\xi$.
We make the choice $\eta = 0$ and adjust $\xi$.

\subsubsection{Steepest descent}

Now, after the discretized and smoothed problem is carefully posed,
we can proceed to solving it by ML.

When finding descent gradients we fix the values of $w_0, b_0, w_2$, and those gradients are supposed to take us to the values of $w_1, b_1$ minimizing the functional

\be
L_{\{Y_s,a_s\}}(w_1, b_1) = \sum_s\left(Y_s-\sum_{j=1}^{M}
\sigma\left(\sum_{i=1}^{M}w_1^{ji}\sigma(a_s-i-1)+b_1^{j}\right)\right)^2
\label{Lid}
\ee
\be
\frac{\partial L_{\{Y_s,a_s\}}}{\partial w_1^{ij}}=0,\ \frac{\partial L_{\{Y_s,a_s\}}}{\partial b_1^{j}}=0
\ee
where $s$ enumerates the set of training samples.

As  already clear from the pictures in the previous subsection,
for certain discretizations, which preserve the true minimum,
the false ones can also occur.
For ML applications to "real life" this is the main source of inspiration,
but for ML of scientific problems this is a difficulty, if not a disaster
on the way to the true answer.

Note also that after discretization
the function $L$ in (\ref{Lid}) is no longer identical zero for
the choice (\ref{wbchoice}) -- it vanishes only for particular (discrete) values of ${a_s}$.
Thus even if we take some pares $(Y,a)=(a,a)$ in the training sequence it is not
guaranteed that they minimize $L$.
We deviate from the true answer (\ref{heavanswer}) not only because of improper $w_i$ and $b_i$,
but also because of the possible (actually, unavoidable) inconsistency/discrepancy
between the choice of $L$ and the training sequences.
This should be taken into account in the process of training.

\subsubsection{Diagonals
}

Since
\be\int dx\,\s\left(\int dy\,w(x,y)\s(a-y)+b(x)\right)=\int dx\,\s\left(\int dy\,\delta(x-y)\s(a-y)+0\right)=\int dx\,\s(a-x)\equiv a
\label{11int}
\ee
one variant of solution is close (in discrete approximation) to or exactly equals (in continuous limit) to
\be
w_1^{ji} = \delta_{ji}=\delta(i - j),\ b_1^j = 0
\ee

The question is whether there  are non-diagonal variants, suggested by TensorFlow computations.

Lets look for an example of "non-diagonal" transformation of $w(x,y), b(x)$ in (\ref{11int}) preserving the equality of left- and right-hand sides in the form:

\be
\int dx\,\s\left(\int dy\,\delta(x-y-c)\s(a-y)+b(x|c)\right)=\int dx\,\s\left(\s(a+c-x)+b(x|c)\right)
\ee
Then we can see that for
\be
b(x|c)=-\chi_{[0,c]}=\left\{
\begin{array}{rl}
-1 & 0\le x\le  c\\
0 & \text{otherwise}
\end{array}
\right.
\ee
we have the equality

\be
&\int dx\,\s\left(\int dy\,\delta(x-y-c)\s(a-y)+b(x|c)\right)=\int dx\,\s\left(\s(a+c-x)-\chi_{[0,c]}\right)=
\nn \\
&= a = \int dx\,\s\left(\int dy\,\delta(x-y)\s(a-y)\right)
\ee

Further, substitute the training sample
$(a,y) = (a,a)$ with $0<a<M$.

No dynamics, nothing happens -- stable point.

\subsubsection{$M=1$
}

For $M=1$ we want to obtain the statement $y=\theta\Big(\theta(a)\Big)$ for the single pair $(a,y)=(1,1)$.

Instead TensorFlow operates with the more complicated $6$-parametric expression
\be
y(a) = w_2\sigma\Big(w_1\sigma(w_0 a + b_0)+b_1\Big) + b_2
\ee
Ideally we should train it with the single pair $(a,y)=(1,1)$,
and clearly there can be not enough data to fix $6$ parameters unambiguously.

\subsubsection{$M=6$
}

Take $M=6$.

For 10 training samples being less then the number of network trainable parameters (being $(3+3) + (3\cdot 3 + 3)+3=21$) after training we get (Fig.9) pretty good samples matching and the structure of $w,b$ remains in the pretty well correspondence with the ansatz (\ref{det11disc}).

While for 40 training samples the configuration becomes unstable and TensorFlow tries to compensate the lack of parameters by the drift of values of $w_1$ (being $3\times 3$ matrix) moving away from initial diagonal matrix (Fig.10).

\begin{figure}
\begin{center}
\includegraphics[scale=0.35]{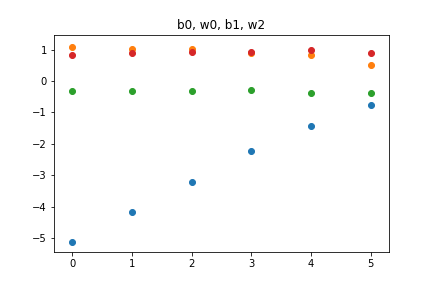}
\includegraphics[scale=0.35]{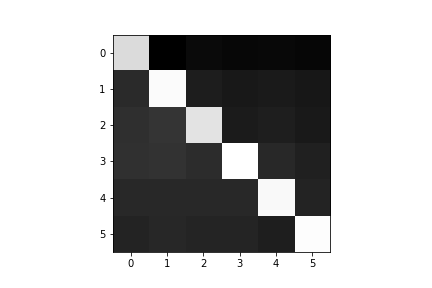}
\includegraphics[scale=0.35]{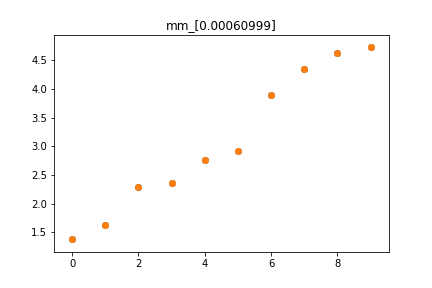}
\caption{Training with $10$ samples. Left: values of $b_0, w_0, b_2, w_2$, middle: values of $w_1$ (close to diagonal), right: matching training sample (orange $y(a)$ completely overlap blue original values of $a$)}
\end{center}
\label{fig:11_10}
\end{figure}

\begin{figure}
\begin{center}
\includegraphics[scale=0.35]{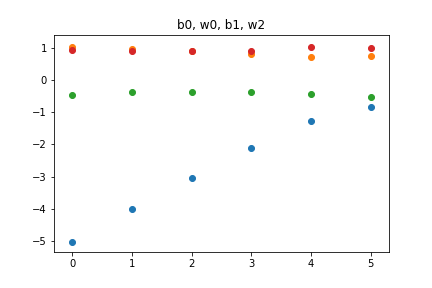}
\includegraphics[scale=0.35]{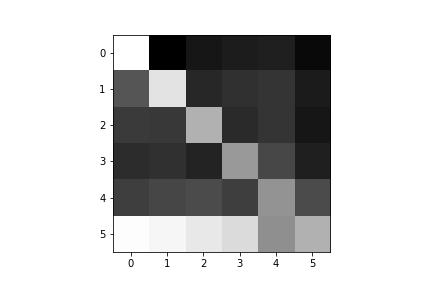}
\includegraphics[scale=0.35]{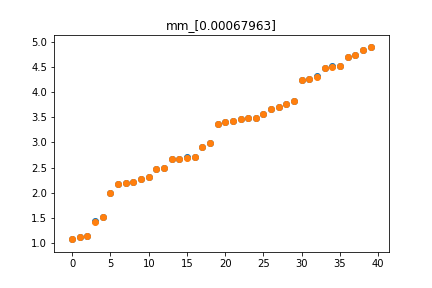}
\caption{Training with $40$ samples. Left: values of $b_0, w_0, b_2, w_2$ (practically the same as for $10$ samples), middle: values of $w_1$ (essentially moved away from diagonal), right: matching training sample (orange $y(a)$ completely overlap blue original values of $a$)}
\end{center}
\label{fig:11_10}
\end{figure}

\subsection{Polynomials}
Heavisidization of a monomial $x_1^{d_1}\dots x_k^{d_k}$ of $k$ variables is obtained from the basic expression (\ref{mult}) (where $d_1+\dots+d_k=n$) by identifying $d$-tuples of variables

$$x_1:=a_1=\dots=a_{d_1},\ \dots\ ,x_k:=a_{n-d_k+1}=\dots=a_n$$
as the sum of terms
\be
(\s(x_1-s_{11}) + \dots +\s(x_1-s_{1d_1}))+\dots + (\s(x_k-s_{k1}) + \dots +\s(x_k-s_{kd_k}))-(\sum_i d_i - 1)
\ee
over the interval of "step" variables $s_{ij}$, i.e.
\begin{align}
x_1^{d_1}\dots x_k^{d_k} \simeq \sum_{\{s_{ij}\}=0}^M\s\left(\sum_{i=1}^n \left(\sum_{j=1}^{k_i} \s(x_i - s_{ij})-d_i\right)+1\right)
\label{mono_eqv}
\end{align}
Then polynomial is a linear combination of expressions (\ref{mono_eqv}) with the weights $w_{d_1\dots d_k}$ being polynomial coefficients.

\be
\sum_{d_1\dots d_k}c_{d_1\dots d_k}x_1^{d_1}\dots x_k^{d_k} \simeq \sum_{d_1\dots d_k}c_{d_1\dots d_k}\sum_{\{s_{ij}\}=0}^M \s\left(\sum_{i=1}^n \left(\sum_{j=1}^{k_i} \s(x_i - s_{ij})-d_i\right)+1\right)
\label{poly_eqv}
\ee

\subsubsection{Example $x^2+3x$}
For the 1st monomial we have in (\ref{mono_eqv}) $i=1, j=1,2$, so
\be
x^2+3x\simeq \sum_{s_1,s_2=0}^M\s\left((\s(x-s_1)+\s(x-s_2) - 2)+1\right)+3\sum_{s=0}^M\s(\s(x-s))=\\
=\sum_{s=0}^M\s(2\s(x-s)-1)+\sum_{0\le s_1<s_2\le M}2\s(\s(x-s_1)+\s(x-s_2)-1)+3\sum_{s=0}^M\s(\s(x-s))
\ee


\subsubsection{The role of ansatz}

On Figs.11,12  illustrated is the process of training a $3\times 3$-determinant model, where the training input was in the space of matrix coefficients and the target values were the values of the corresponding determinant. On Fig.12 we show the training progress in case when initial weights were taken as random, while on Fig.11 there is the progress for taking initial weights according to determinant Hevisidization ansatz (\ref{hdet}). Initializing network parameters via ansatz results in decrease of loss function to a relatively low value, while the parameters/weights undergo just a minor correction. Starting from random initialization results in pretty poor (at least slow) training. These checks allow us to conclude that using the above polynomial formulas for network weights gives us the states being initially pretty close to optimal.

\begin{figure}[h]
\begin{center}
\includegraphics[scale=0.5]{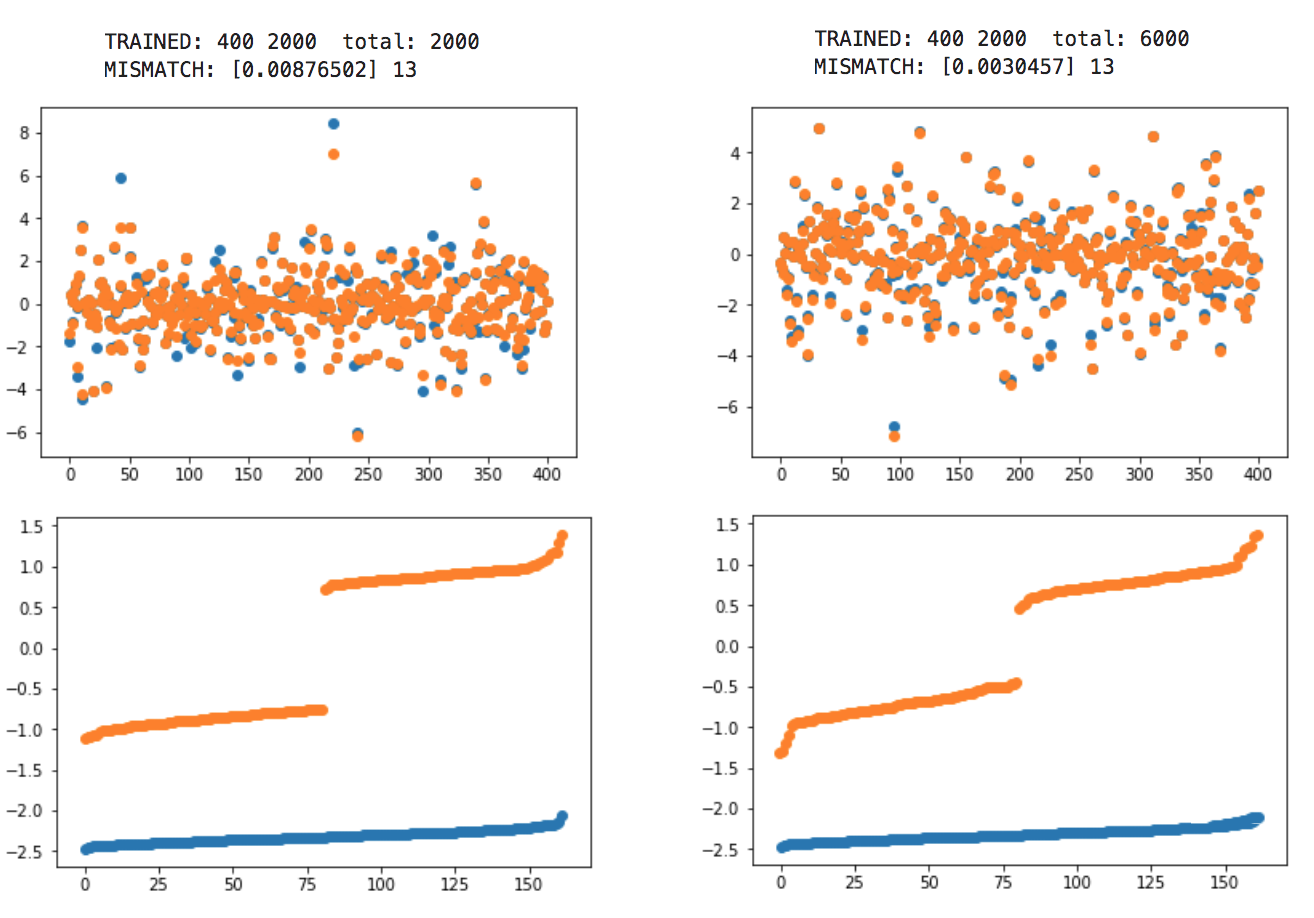}
\caption{Evolution of $3\times 3$-determinant model output (top) and some learnable parameters (bottom) after 2000 and 6000 training steps correspondingly: the case of initializing parameters according to {\bf determinant ansatz} - essential decreasing loss values, minor corrections of network weights}
\end{center}
\label{fig:hint}
\end{figure}

\begin{figure}[H]
\begin{center}
\includegraphics[scale=0.5]{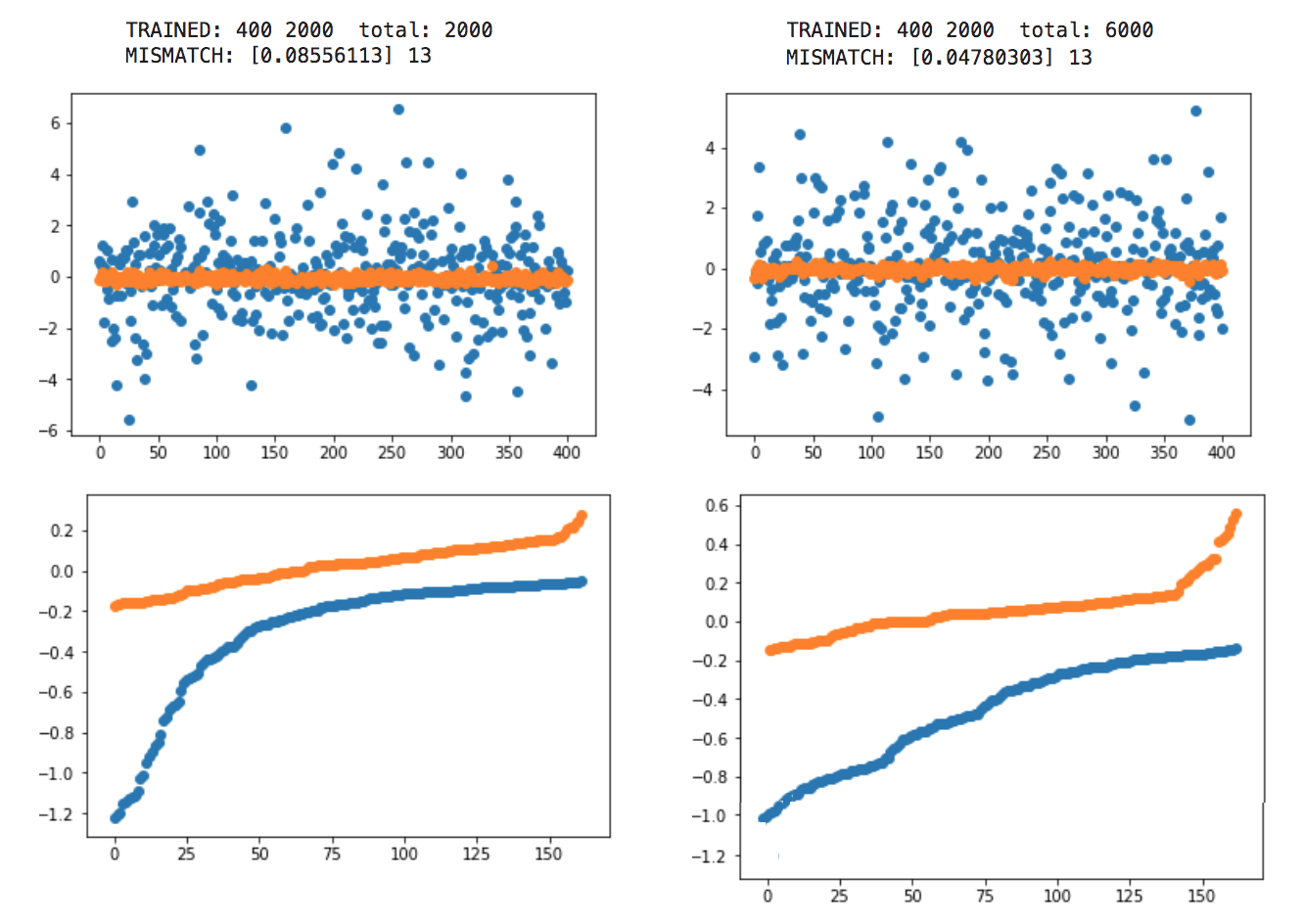}
\caption{Evolution of $3\times 3$-determinant model output (top) and some learnable parameters (bottom) after 2000 and 6000 training steps correspondingly: the case of {\bf randomly initialized} parameters - high remaining loss value, active changes of network weights}
\end{center}
\label{fig:no_hint}
\end{figure}

\section*{}

\hide{
\begin{figure}
\begin{center}
\includegraphics[scale=0.5]{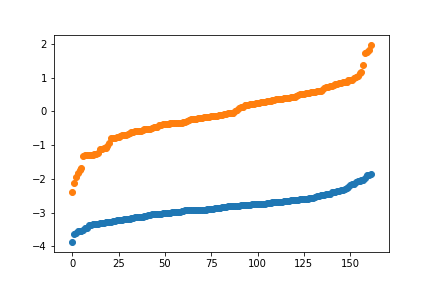}
\caption{ }
\end{center}
\label{fig:hint_lim}
\end{figure}

\begin{figure}
\begin{center}
\includegraphics[scale=0.5]{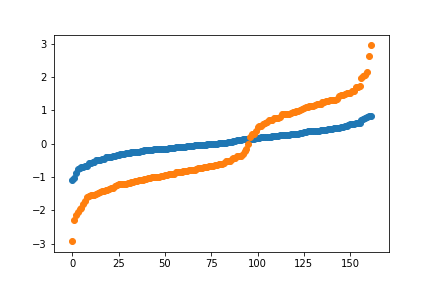}
\caption{ }
\end{center}
\label{fig:nohint_lim}
\end{figure}

\begin{figure}
\begin{center}
\includegraphics[scale=0.5]{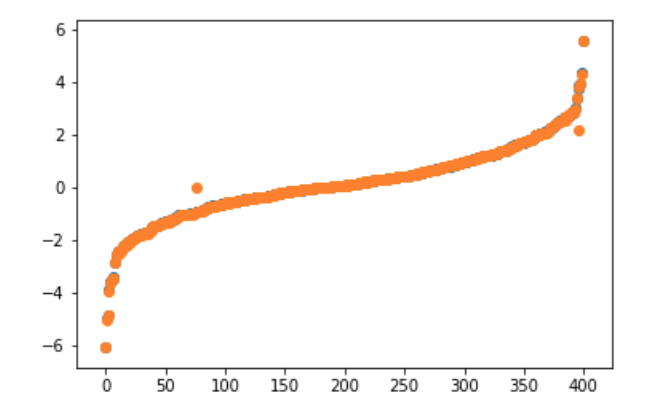}
\caption{ }
\end{center}
\label{fig:sample}
\end{figure}

\begin{figure}
\begin{center}
\includegraphics[scale=0.5]{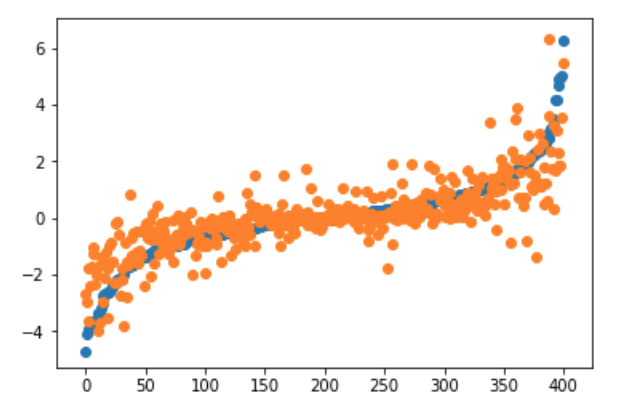}
\caption{ }
\end{center}
\label{fig:test}
\end{figure}

\begin{figure}
\begin{center}
\includegraphics[scale=0.3]{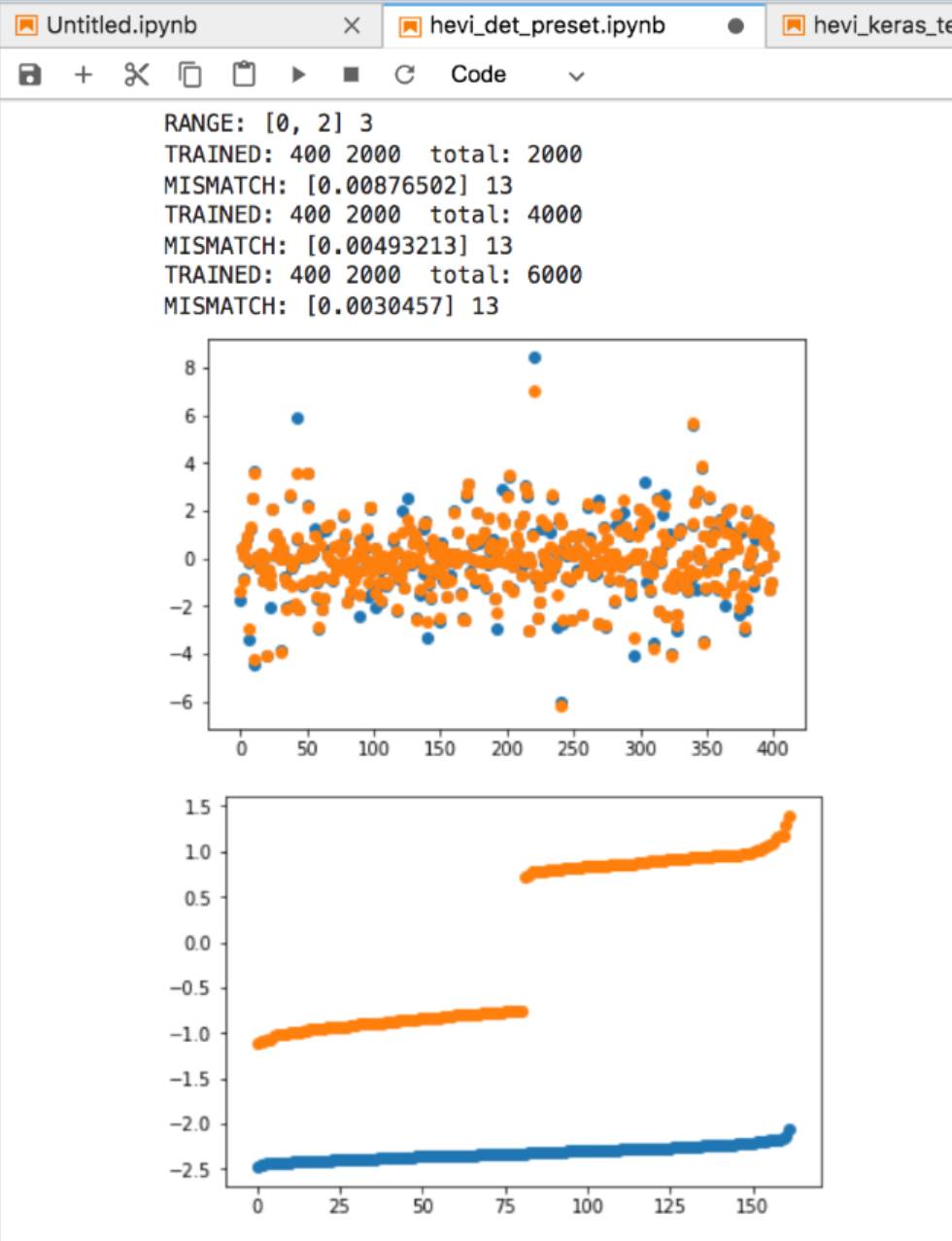}
\caption{ }
\end{center}
\label{fig:hint}
\end{figure}

\begin{figure}
\begin{center}
\includegraphics[scale=0.5]{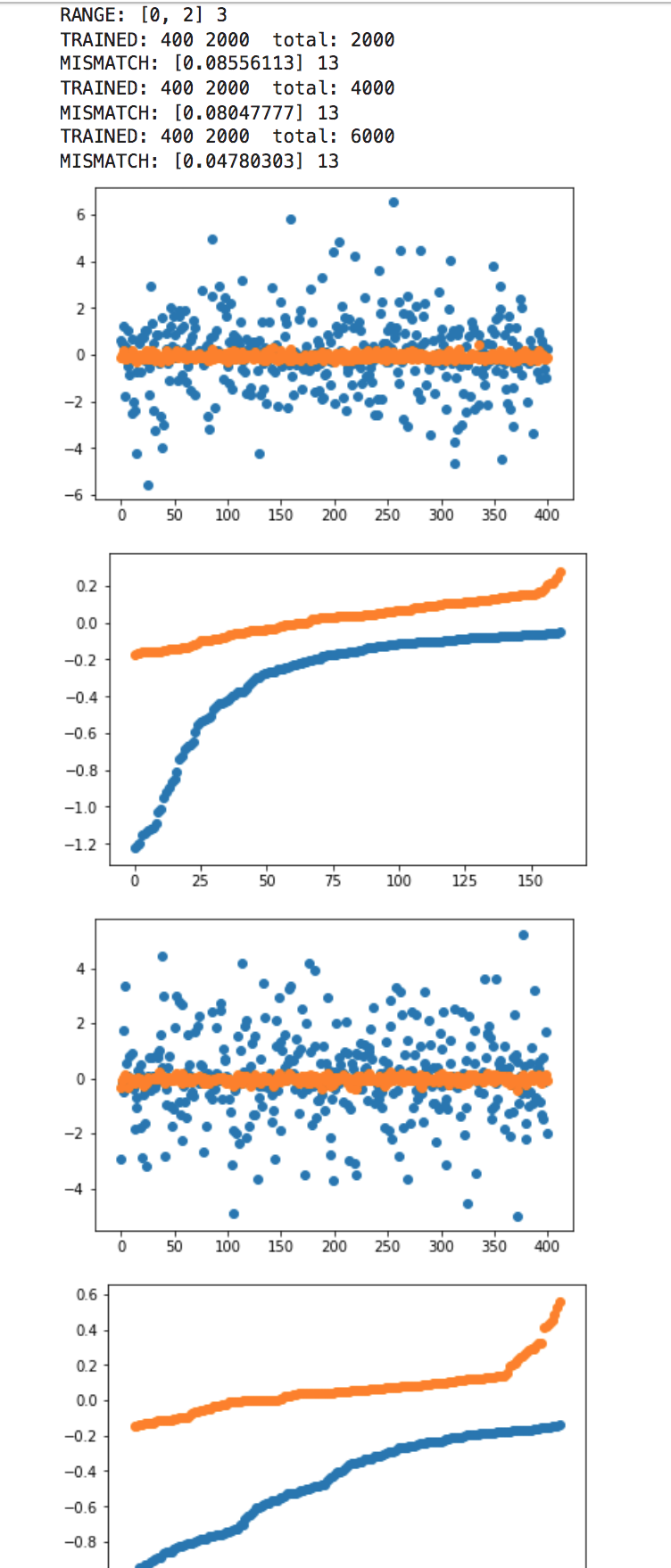}
\caption{ }
\end{center}
\label{fig:no_hint}
\end{figure}
}

\section{Heaviside transform}
One-layer network with 1-dimensional input $x$

\be
{\cal I}(x)=\sum_kw(k)\s(x + k)
\label{h_transform}
\ee
may be considered as an expansion of $y(x)$ into an analogue of Fourier series, with $\s(x+k)$ playing the role of harmonics. Then the steepest descent solution for coefficients $w(k)$ is the inverse Heaviside transform.

The fact that in reality we always have only limited number of terms in (\ref{h_transform}) unavoidably makes the problem of finding $w(k)$ for infinite range of output ${\cal I}$ not well posed, forcing TensorFlow infinitely wander between degenerate states none of which may satisfy all of the training samples.

\section{Discussion
\label{approx}
}

For a (continouos) function representable point-wise by Fourier integral $y(x)=\int a_\omega e^{i\omega x}d\omega$ lets modify the "representation"\ task: suppose we are allowed to use only a finite set of $N$ harmonics, then
\vskip 3mm
{\it what will be the best approximation of $y(x)$ by a set of $N$ harmonics $Y(x) = \sum_{k=1}^N a_ke^{i\omega_kx}$?}
\vskip 3mm

This immediately turns the exact Fourier transform task into a variational problem with respect to the values of $\omega$ and $a$, with a loss functional $L = \|y - Y\|$. But the solution of it will still enjoy the fundamental property of Fourier transform, the uncertainty principle:

\vskip 3mm

{\it sharp changes in the function $y$ result in frequencies $\omega$ spreading}
\vskip 3mm

Going to multi-dimensional spaces for $x$ and $y$, we can rewrite our Fourier approximation expression $Y(\vec x)$ as:

$$y_j = \sum_{k=1}^NA^k_js(\omega^i_kx_i)$$
where $\omega$ is a linear $\dim x\times N$ operator, $A$ is a linear $N\times\dim y$ operator, $s:=\sin$, and in the affine chart we can always take $x_0\equiv 1$, which is equivalent to using the phase ("bias"\ in network terms).

In a similar way, for a 1-layer network $\vec x\mapsto Y(\vec x)=w_1\sigma(w_0x)$ we can write:

$$y_j = \sum_{k=1}^N w_{1,j}^k \sigma(w^i_{0,k}x_i)$$
\noindent
where $N$ is the number of nodes in our network, $w_0$ and $w_1$ are linear $\dim x\times N$ and $N\times\dim y$ operators correspondingly, and instead of our standard $\sin$ function we have an activation function $\sigma$, taken above to be sigmoid. Here the use of bias $b$ is replaced by going to homogeneous coordinates where $\sum_{i=1}w^ix_i+b$ becomes a linear expression $\sum_{i=0}w^ix_i$, which returns to bias-using variant when setting $x_0\equiv 1$.

This is readily extendable to the case of more layers (which may be actually done for Fourier approximation as well):
$$Y(\vec x) := w_n\sigma\circ\dots\circ w_1\sigma(w_0x)$$
\noindent
where each $w_l$ is a linear $N_l\times N_{l+1}$ operator between layers.

The solution of the corresponding variational problem may be called a {\bf ($\sigma$-)transform/mapping} from the space of functions $y$ to the space of approximation coefficients $w_l$ - weights of our network. We foresee that for each activation function $\sigma$ there is an analogue of uncertainty principle, i.e. a statement (maybe just qualitative) relating certain properties of functions $y$ to some properties of the sets of approximating coefficients $w$.

In the case of $\sigma$ being sigmoid and 1-layer network, based on the observations described in the previous sections we formulate the "uncertainty principle"\ as:
\vskip 4mm
{\it The more "oscillating/wavy/irregular"\ is the target function $y$ on a given interval, the wider is the spectrum (the set of relatively large coefficients of $w_1$) of its $\sigma$-transform.}
\vskip 4mm

This formulation certainly needs a quantitative criterion (as an example, $y$ being a single harmonic having maximally narrow Fourier-approximation spectrum, is maximally "irregular"\ function in terms of $\sigma$-approxi\-mations). But even in this form it has practical implications for the networks training. Suppose for a function with given "regularity"\ we found a $\sigma$-approximation having required precision, so all the components of our $\sigma$-transform, or nodes of our network, have made a proper non-zero impact into approximating expression $Y(\vec x)$. Now trying to add more nodes to the layer will result in having "duplicate components"\,, i.e. we will increase dimensionality of our variational parameters space $w$ creating there subspaces with practically constant values of the functional (approximation precision). And when our gradient descent trajectory gets to such subspace (called "canyons"\  above) the speed of descent (thus, of the training) will drop down dramatically. Which implies that for a given "degree of regularity"\ of training target values, there is an {\it optimal number} of network elements, and exceeding that number may significantly increase the training time, without actual impact to precision.

As an example of that optimal number of network elements, for a polynomial target function it is claimed to be given by formula (\ref{poly_eqv}).

\vskip 10mm

\section{Conclusion}

Surprisingly or not, phenomena and problems, encountered in ML application to exact sciences,
are themselves pure scientific and belong to physics, not to computer science.
On the other hand, they sound slightly different and shed new light on the
well-known phenomena -- for example, extend the uncertainty principle from Fourier
and, later, wavelet analysis to a new peculiar class of nearly singular
sigmoid functions.

\section*{Acknowledgements}

The work was supported
by the state assignment of the Institute for Information Transmission Problems of RAS.

\end{document}